\documentclass[runningheads]{llncs}
\usepackage[T1]{fontenc}
\usepackage{graphicx}
\usepackage{graphicx}
\usepackage{microtype}
\usepackage{amsmath}
\usepackage{amsfonts}
\usepackage{microtype}
\usepackage{graphicx}
\usepackage{multirow}
\usepackage{booktabs}
\usepackage{hhline}
\usepackage{xspace}
\usepackage{color}
\usepackage{enumitem}
\usepackage{amsmath}
\usepackage{amssymb}
\usepackage{pifont}
\usepackage{amsfonts}
\usepackage{microtype}
\usepackage{graphicx}
\usepackage{multirow}
\usepackage{booktabs}
\usepackage{hhline}
\usepackage{xspace}
\usepackage{color}
\usepackage{enumitem}
\usepackage{ulem}
\usepackage{comment}
\usepackage{bbding}
\usepackage{bm}
\usepackage[ruled,linesnumbered]{algorithm2e}
\usepackage[pagebackref=true,breaklinks=true,colorlinks=true,bookmarks=false,citecolor=blue,urlcolor=black,linkcolor=blue]{hyperref}
\newcommand{\cmark}{\ding{51}}%
\newcommand{\xmark}{\ding{55}}%
\begin{document}
\title{ArSDM: Colonoscopy Images Synthesis with Adaptive Refinement Semantic Diffusion Models}
\titlerunning{ArSDM}
\author{
    Yuhao Du       \inst{1,2\footnotemark[4]} \and       % index{Du, Yuhao}
    Yuncheng Jiang \inst{1,2,3\footnotemark[4]} \and       % index{Jiang, Yuncheng}
    Shuangyi Tan   \inst{1,2} \and       % index{Tan, Shuangyi}
    Xusheng Wu     \inst{6} \and \\      % index{Wu, Xusheng}
    Qi Dou         \inst{5} \and         % index{Dou, Qi}
    Zhen Li        \inst{2,3\footnotemark[1]} \and        % index{Li, Zhen}
    Guanbin Li     \inst{4\footnotemark[1]} \and    % index{Li, Guanbin}
    Xiang Wan      \inst{1,2}            % index{Wan, Xiang}
    }
\authorrunning{Y. Du et al.}
\institute{Shenzhen Research Institute of Big Data, Shenzhen, China \\ \and
    SSE, CUHK-Shenzhen, Shenzhen, China \\ \and
    FNii, CUHK-Shenzhen, Shenzhen, China \\ \and
    School of Computer Science and Engineering, Research Institute of Sun Yat-sen University in Shenzhen, Sun Yat-sen University, Guangzhou, China \\ \and
    The Chinese University of Hong Kong, Hong Kong, China \\ \and
    Shenzhen Health Development Research and Data Management Center, China \\
    \email{lizhen@cuhk.edu.cn} \\
    \email{liguanbin@mail.sysu.edu.cn}}
\maketitle              % typeset the header of the contribution
\renewcommand{\thefootnote}{\fnsymbol{footnote}}
\footnotetext[4]{Equal contributions.}
\footnotetext[1]{Corresponding authors.}
\begin{abstract}
Colonoscopy analysis, particularly automatic polyp segmentation and detection, is essential for assisting clinical diagnosis and treatment. However, as medical image annotation is labour- and resource-intensive, the scarcity of annotated data limits the effectiveness and generalization of existing methods. Although recent research has focused on data generation and augmentation to address this issue, the quality of the generated data remains a challenge, which limits the contribution to the performance of subsequent tasks. Inspired by the superiority of diffusion models in fitting data distributions and generating high-quality data, in this paper, we propose an \textbf{A}daptive \textbf{R}efinement \textbf{S}emantic \textbf{D}iffusion \textbf{M}odel (\textbf{ArSDM}) to generate colonoscopy images that benefit the downstream tasks. Specifically, ArSDM utilizes the ground-truth segmentation mask as a prior condition during training and adjusts the diffusion loss for each input according to the polyp/background size ratio. Furthermore, ArSDM incorporates a pre-trained segmentation model to refine the training process by reducing the difference between the ground-truth mask and the prediction mask. Extensive experiments on segmentation and detection tasks demonstrate the generated data by ArSDM could significantly boost the performance of baseline methods.
% The source code is available at~\url{https://github.com/DuYooho/ArSDM}.

\end{abstract}
%
%
%

% ****************** Section 1 ******************
\section{Introduction}
% ****************** Figure 1 ******************
\begin{figure}[t]
\centering
\includegraphics[width=\textwidth]{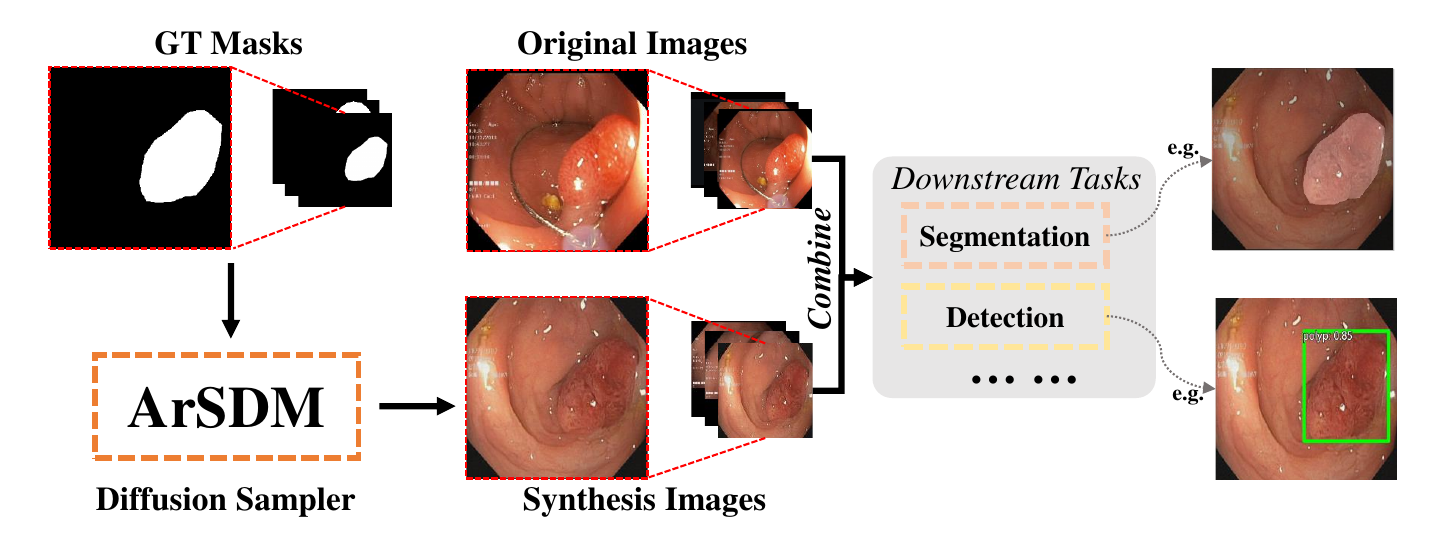}
\caption{Overview of the pipeline of our proposed approach, where details of \textbf{ArSDM} are described in section~\ref{sec:method}.} 
\label{fig:fig1}
\end{figure}
% ****************** Figure 1 ******************
Colonoscopy is a critical tool for identifying adenomatous polyps and reducing rectal cancer mortality. Deep learning methods have shown powerful abilities in automatic colonoscopy analysis, including polyp segmentation \cite{fan2020pranet,wei2021shallow,zhang2022lesion,zhang2020adaptive,zhao2022semi} and polyp detection~\cite{NimaTajbakhsh2016AutomatedPD,LingyunWu2021MultiframeCF}. However, the scarcity of annotated data due to high manual annotation costs results in poorly trained and low generalizable models. Previous methods have relied on generative adversarial networks (GANs)~\cite{ma2020cycle,xu2020ofgan} or data augmentation methods~\cite{chaitanya2021semi,sandfort2019data,zhao2019data} to enhance learning features, but these methods yielded limited improvements in downstream tasks. Recently, diffusion models \cite{ho2020denoising,sohl2015deep} have emerged as promising solutions to this problem, demonstrating remarkable progress in generating multiple modalities of medical data~\cite{dhariwal2021diffusion,park2019semantic,rombach2022high,wang2022semantic}.

Despite recent progress in these methods for medical image analysis, existing models face two major challenges when applied to colonoscopy image analysis. Firstly, the foreground (polyp) of colonoscopy images contains rich pathological information yet is often tiny compared with the background (intestine wall) and can be easily overwhelmed during training. Thus, naive generative models may generate realistic colonoscopy images but those images seldom contain polyp regions. In addition, in order to generate high-quality annotated samples, it is crucial to maintain the consistency between the polyp morphologies in synthesized images and the original masks, which current generative models struggle to achieve.

To tackle these issues and inspired by the remarkable success achieved by diffusion models in generating high-quality CT or MRI data~\cite{kim2022diffusion,pinaya2022fast,wolleb2022diffusion}, we creatively propose an effective adaptive refinement semantic diffusion model (ArSDM) to generate polyp-contained colonoscopy images while preserving the original annotations. The pipeline of the data generation and downstream task training is shown in Fig.~\ref{fig:fig1}. Specifically, we use the original segmentation masks as conditions to train a conditional diffusion model, which makes the generated samples share the same masks with the input images. Moreover, during diffusion model training, we employ an adaptive loss re-weighting method to assign loss weights for each input according to the size ratio of polyps and background, which addresses the overfitting problem for the large background. In addition, we fine-tune the diffusion model by minimizing the distance between the original ground truth masks and the prediction masks from synthesis images via a pre-trained segmentation network. Thus the refined model could generate samples better aligned with the original masks.

In summary, our contributions are three-fold: (1) \textbf{Adaptive Refinement SDM}: Based on the standard semantic diffusion model \cite{wang2022semantic}, we propose a novel ArSDM with the adaptive loss re-weighting and the prediction-guided sample refinement mechanisms, which is capable of generating realistic polyp-contained colonoscopy images while preserving the original annotations.
To the best of our knowledge, this is the first work for adapting diffusion models to colonoscopy image synthesis. (2) \textbf{Large-Scale Colonoscopy Generation}: The proposed approach can be used to generate large-scale datasets with no/arbitrary annotations, which significantly benefits the medical image society, laying the foundation for large-scale pre-training models in automatic colonoscopy analysis. (3) \textbf{Qualitative and Quantitative Evaluation}: We conduct extensive experiments to evaluate our method on five public benchmarks for polyp segmentation and detection. The results demonstrate that our approach could help deep learning methods achieve better performances.
The source code is available at~\url{https://github.com/DuYooho/ArSDM}.

% ****************** Section 2 ******************
\section{Method}
\label{sec:method}
\noindent\textbf{Background}~
Denoising diffusion probabilistic models (DDPMs) \cite{ho2020denoising} are classes of deep generative models, which have forward and reverse processes. 
The forward process is a Markov Chain that gradually adds Gaussian noise to the original data.
This process can be formulated as the joint distribution $q\left(\mathbf{x}_{1: T} \mid \mathbf{x}_0\right)$:
\begin{equation}
q\left(\mathbf{x}_{1: T} \mid \mathbf{x}_0\right):=\prod_{t=1}^T q\left(\mathbf{x}_t \mid \mathbf{x}_{t-1}\right),
q\left(\mathbf{x}_t \mid \mathbf{x}_{t-1}\right):=\mathcal{N}\left(\mathbf{x}_t ; \sqrt{1-\beta_t} \mathbf{x}_{t-1}, \beta_t \mathbf{I}\right),
\end{equation}
where $q\left(\mathbf{x}_{0}\right)$ is the original data distribution with $\mathbf{x}_{0} \sim q\left(\mathbf{x}_{0}\right)$,
$\mathbf{x}_{1: T}$ are latents with the same dimension of $\mathbf{x}_0$
and $\beta_t$ is a variance schedule.

The reverse process is aiming to learn a model to reverse the forward process that reconstructs the original input data,
which is defined as:
\begin{equation}
p_\theta\left(\mathbf{x}_{0: T}\right):=p\left(\mathbf{x}_T\right) \prod_{t=1}^T p_\theta\left(\mathbf{x}_{t-1} \mid \mathbf{x}_t\right),
p_\theta\left(\mathbf{x}_{t-1} \mid \mathbf{x}_t\right):=\mathcal{N}\left(\mathbf{x}_{t-1} ; \bm{\mu}_\theta\left(\mathbf{x}_t, t\right), \sigma^2_t \mathbf{I} \right)
\label{eq:reverse},
\end{equation}
where $p\left(\mathbf{x}_T\right)$ is the noised Gaussian transition from the forward process at timestep $T$.
In this case, we only need to use deep-learning models to represent $\bm{\mu}_\theta$ with $\theta$ as the model parameters.
According to the original paper \cite{ho2020denoising}, the loss function can be simplified as:
\begin{equation}
\mathcal{L}_{\text {simple }} := \mathbb{E}_{t, \mathbf{x}_t, \bm{\epsilon}  \sim \mathcal{N} \left( \mathbf{0},\mathbf{I} \right)}\left[\left\|\bm{\epsilon}-\bm{\epsilon}_\theta\left( \mathbf{x}_t, t \right)\right\|^2 \right].
\label{eq:5}
\end{equation}
Thus, instead of training the model $\bm{\mu}_\theta$ to predict $\tilde{\bm{\mu}}_t$, we can train the model $\bm{\epsilon}_\theta$ to predict $\tilde{\bm{\epsilon}}$, which is easier for parameterization and learning.

\begin{figure}[t]
    \centering
    \includegraphics[width=\textwidth]{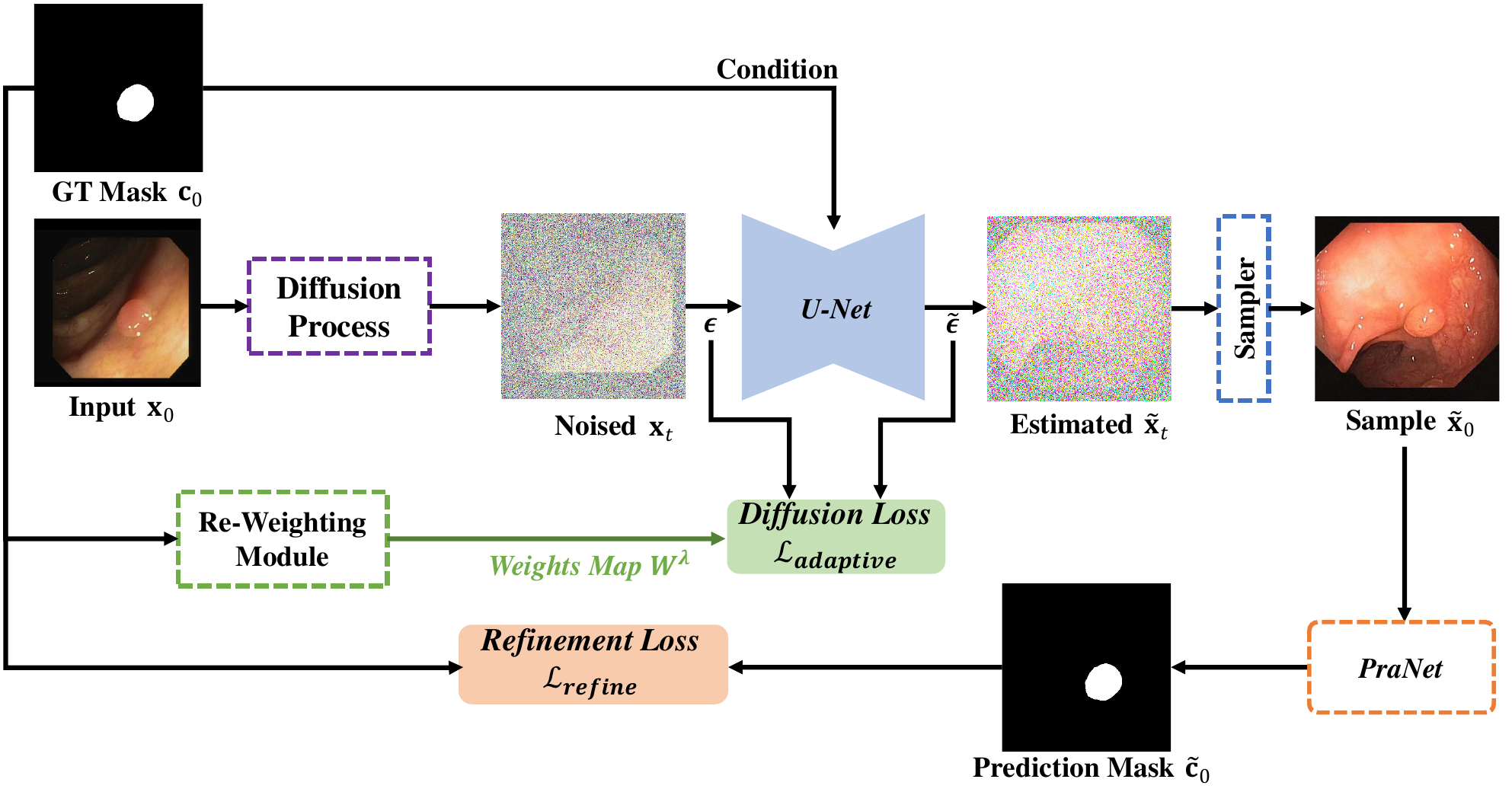}
    \caption{The overall architecture of \textbf{ArSDM}.} 
    \label{fig:framework}
\end{figure}

In this paper, we propose an adaptive refinement semantic diffusion model,
a variant of DDPM,
which has three key parts, \textit{i.e.}, mask conditioning, adaptive loss re-weighting, and prediction-guided sample refinement. The overall illustration of our framework is shown in Fig.~\ref{fig:framework}.

% ****************** Figure 2 ******************
% ****************** Figure 2 ******************

\subsection{Mask Conditioning}
Unlike the previous generative methods, our work aims to generate a synthetic image with an identical segmentation mask to the original annotation. 
To accomplish this, we adapt the widely used conditional U-Net architecture \cite{wang2022semantic} in the reverse process,
where the mask is fed as a condition.
Specifically, for an input image $\mathbf{x}_0 \in \mathbb{R}^{H \times W \times C}$,
$\mathbf{x}_t$ can be sampled at any timestep $t$ with the closed form:
\begin{equation}
\mathbf{x}_t = \sqrt{\bar{\alpha}_t} \mathbf{x}_0+\sqrt{1-\bar{\alpha}_t} \bm{\epsilon},
\end{equation}
where $\bm{\epsilon} \sim \mathcal{N} \left( \mathbf{0},\mathbf{I} \right), \alpha_t:=1-\beta_t$ and $\bar{\alpha}_t:=\prod_{s=1}^t \alpha_s$.
It will be fed into the encoder $\mathcal{E}$ of the U-Net,
and its corresponding mask annotation $\mathbf{c}_0 \in \mathbb{R}^{H \times W}$ will be injected into the decoder $\mathcal{D}$.
The model output can be formulated as:
\begin{equation}
\bm{\epsilon}_\theta\left( \mathbf{x}_t, t, \mathbf{c}_0 \right) = \mathcal{D} \left( \mathcal{E} \left( \mathbf{x}_t \right), \mathbf{c}_0 \right).
\end{equation}
Thus, the U-Net model $\bm{\epsilon}_\theta$ in Eq. \ref{eq:5} becomes $\bm{\epsilon}_\theta \left( \mathbf{x}_t, t, \mathbf{c}_0 \right)$,
and the loss function in Eq.~\ref{eq:5} is changed to:
\begin{equation}
\mathcal{L}_{\text {condition }}=\mathbb{E}_{t, \mathbf{x}_t, \mathbf{c}_0, \bm{\epsilon} \sim \mathcal{N} \left( \mathbf{0},\mathbf{I} \right)}\left[\left\|\bm{\epsilon}-\bm{\epsilon}_\theta\left( \mathbf{x}_t, t, \mathbf{c}_0 \right)\right\|^2 \right].
\end{equation}

\subsection{Adaptive Loss Re-Weighting}
The polyp regions in the colonoscopy images differ from the background regions, which contain more pathological information and should be adequately treated to learn a better model.
However, training the diffusion models using the original loss function ignores the difference between different regions, where each pixel shares the same weights when calculating the loss.
This would lead to the model generating more background-like polyps since the large background region will easily overwhelm the small foreground polyp regions during training.
A simple way to alleviate this problem is to apply a weighted loss function that assigns the polyp and background regions with different weights.
However, most polyps vary a lot in size and shape. 
Thus assigning constant weights for all polyps exacerbated the imbalance problem.
In this case, to tackle this problem, we propose an adaptive loss function that vests different weights according to the size ratio of the polyp over the background.
Specifically, we define a pixel-wise weights matrix $W^{\lambda} \in \mathbb{R}^{H \times W}$
with each entry $w^{\lambda}_{i,j}$ to be:
\begin{equation}
w^{\lambda}_{i,j}=
\begin{cases}
1 - r& \text{, } p=1\\
r& \text{, } p=0
\end{cases},\qquad
r = \frac{\#(p=1)}{H \times W},
\end{equation}
where $p=1$ means the pixel $p$ at $(h,w)$ belongs to the polyp region and $p=0$ means it belongs to the background region.
Thus, the loss function becomes:
\begin{equation}
\mathcal{L}_{\text {adaptive }}=\mathbb{E}_{t, \mathbf{x}_t, \mathbf{c}_0, \bm{\epsilon} \sim \mathcal{N} \left( \mathbf{0},\mathbf{I} \right)}\left[ 
W^{\lambda} \cdot \left\|\bm{\epsilon}-\bm{\epsilon}_\theta\left( \mathbf{x}_t, t, \mathbf{c}_0 \right)\right\|^2 \right].
\end{equation}

% ******************* Algorithm 1 **********************
\begin{algorithm}[t]
\caption{One training iteration of ArSDM}
    \label{alg:algorithm}
    \KwIn{$t \sim $ Uniform($\{1,...,T\}$), $\mathbf{x}_0 \sim q(\mathbf{x}_0)$, $\mathbf{c}_0$, $\bm{\epsilon} \sim \mathcal{N} \left( \mathbf{0},\mathbf{I} \right)$}
    \KwOut{$\tilde{\bm{\epsilon}}$, $\tilde{\mathbf{c}}_0$}
    $\mathbf{x}_t = \sqrt{\bar{\alpha}_t} \mathbf{x}_0+\sqrt{1-\bar{\alpha}_t} \bm{\epsilon}$;\quad
    $\tilde{\mathbf{x}}_t = \sqrt{\bar{\alpha}_t} \mathbf{x}_0+\sqrt{1-\bar{\alpha}_t} \bm{\epsilon}_\theta\left( \mathbf{x}_t, t, \mathbf{c}_0 \right)$\\
    \textbf{for} $i=t,...,1$ \textbf{do} \\
    \quad $\mathbf{z} \sim \mathcal{N}(\mathbf{0}, \mathbf{I})$ if $i>1$, else $\mathbf{z}=\mathbf{0}$;\quad
    $\tilde{\mathbf{x}}_{i-1}=\frac{1}{\sqrt{\alpha_i}}\left(\tilde{\mathbf{x}}_i-\frac{1-\alpha_i}{\sqrt{1-\bar{\alpha}_i}} \bm{\epsilon}_\theta\left(\tilde{\mathbf{x}}_i, i, \mathbf{c}_0\right)\right)+\sigma_i \mathbf{z}$ \\
    \textbf{end for} \\
    $\tilde{\mathbf{c}}_0 = \mathcal{P}(\tilde{\mathbf{x}}_0)$ \\
    Take gradient descent step on $\nabla_\theta \mathcal{L}_{\text {total }}$ 
\end{algorithm}
% ******************* Algorithm 1 **********************

\subsection{Prediction-Guided Sample Refinement}
The downstream tasks of polyp segmentation and detection require rich semantic information on polyp regions to train a good model.
Through extensive experiments, we found inaccurate sample images with coarse polyp boundary that is not aligned properly with the original masks may introduce large biases and noises to the datasets.
The model can be confused by several conflicting training images with the same annotation.
To this end, we design a refinement strategy that uses the prediction of a pre-trained segmentation model on the sampled images to guide the training process and restore the proper polyp boundary information.
Specifically, at each iteration of training, the output $\tilde{\bm{\epsilon}} = \bm{\epsilon}_\theta\left( \mathbf{x}_t, t, \mathbf{c}_0 \right)$ will go into the sampler to generate sample image $\tilde{\mathbf{x}}_0$.
Then, we take the sample image as the input of the segmentation model to predict the pseudo masks $\tilde{\mathbf{c}}_0$. 
We propose the following refinement loss based on IoU loss and binary cross entropy (BCE) loss between $\tilde{\mathbf{c}}_0$ and $\mathbf{c}_0$.
The refinement loss is:
\begin{equation}
\begin{aligned}
\mathcal{L}_{\text {refine }}  =   
\mathcal{L} (\mathbf{c}, \tilde{\mathbf{c}_g}) + \sum_{i=3}^{i=5} \mathcal{L}\left( \tilde{\mathbf{c}_i} \right), \\
\tilde{\mathbf{c}}_0 = 
\left \{ \tilde{\mathbf{c}_3}, \tilde{\mathbf{c}_4}, \tilde{\mathbf{c}_5}, \tilde{\mathbf{c}_g}  \right \} =
\mathcal{P} \left( \mathcal{S} \left( \tilde{\bm{\epsilon}} \right) \right), 
\end{aligned}
\end{equation}
where $\mathcal{L} = \mathcal{L}_{I o U}+\mathcal{L}_{B C E}$ is the sum of the IoU loss and BCE loss,
$\tilde{\mathbf{c}}_0$ is the collection of the three side-outputs ($\tilde{\mathbf{c}_3}, \tilde{\mathbf{c}_4}, \tilde{\mathbf{c}_5}$) and the global map $\tilde{\mathbf{c}_g}$ as described in \cite{fan2020pranet}.
$\mathcal{P}(\cdot)$ represents the PraNet model and $\mathcal{S}(\cdot)$ is the DDIM \cite{song2021denoising} sampler.
The detailed procedure of one training iteration is shown in Algorithm~\ref{alg:algorithm} and
the overall loss function is defined as:
\begin{equation}
\mathcal{L}_{\text {total }} = \mathcal{L}_{\text {adaptive }} + \mathcal{L}_{\text {refine }}.
\end{equation}

% ****************** Section 3 ******************
\section{Experiments}
% ****************** Table 1: seg_results ******************
\begin{table*}[t]
\centering
\caption{Comparisons of different settings applied on three polyp segmentation baselines.}
\renewcommand\tabcolsep{0.9pt}
\begin{tabular}{@{}lcccccccccc|cc@{}}
\toprule
\multirow{2}{*}{Methods} 
& \multicolumn{2}{c}{EndoScene}                      
& \multicolumn{2}{c}{ClinicDB}     
& \multicolumn{2}{c}{Kvasir}   
& \multicolumn{2}{c}{ColonDB}   
& \multicolumn{2}{c}{ETIS}   
& \multicolumn{2}{|c}{Overall}  
\\ \cmidrule(l){2-13} 
&mDice  &mIoU  &mDice  &mIoU  &mDice  &mIoU  &mDice  &mIoU  &mDice  &mIoU  &mDice  &mIoU        \\             
\midrule
PraNet          &87.1           &79.7           &89.9           &84.9   &89.8   &84.0   &70.9   &64.0   &62.8   &56.7   &74.0   &67.5    \\    
+LDM            &83.7           &76.9           &88.2           &83.5   &88.4   &83.0   &62.6   &56.0   &56.2   &50.3   &67.8   &61.7    \\
+SDM            &\textbf{89.9}  &\textbf{83.2}  &89.2           &83.7   &88.4   &82.6   &74.2   &66.5   &66.4   &60.3   &76.4   &69.6    \\
+\textbf{Ours}  &89.7           &82.7           &\textbf{93.3}  &\textbf{88.5}   &\textbf{89.9}   &\textbf{84.5}   &\textbf{76.1}   &\textbf{68.9}   &\textbf{75.5}   &\textbf{68.1}   &\textbf{80.0}   &\textbf{73.2}  \\
\midrule
SANet  &88.8   &81.5   &\textbf{91.6}  &85.9   &90.4   &84.7   &75.3   &67.0   &75.0   &65.4   &79.4   &71.4    \\
+LDM   &72.7   &60.5   &88.8           &82.8   &88.7   &82.7   &64.3   &55.4   &58.0   &49.2   &68.3   &59.8    \\
+SDM   &\textbf{90.2}  &83.0   &89.9   &84.1   &90.9   &85.4   &77.6   &69.3   &74.7   &66.8   &80.4   &72.9    \\
+\textbf{Ours}   &\textbf{90.2}   &\textbf{83.2}   &91.4   &\textbf{86.1}   &\textbf{91.1}   &\textbf{85.6}   &\textbf{77.7}   &\textbf{70.0}   &\textbf{78.0}   &\textbf{69.5}   &\textbf{81.5}   &\textbf{74.1} \\
\midrule
PVT    &\textbf{90.0}   &\textbf{83.3}   &93.7   &88.9   &\textbf{91.7}   &\textbf{86.4}   &80.8   &72.7   &78.7   &70.6   &83.3   &76.0    \\
+LDM   &88.2   &81.2   &92.3   &87.1   &91.2   &85.7   &78.7   &70.4   &78.0   &69.6   &81.9   &74.2    \\
+SDM   &88.8   &81.7   &\textbf{93.9}   &\textbf{89.2}   &91.2   &86.1   &81.3   &73.5   &78.7   &71.1   &83.4   &76.3    \\
+\textbf{Ours}   &88.2   &81.2   &92.2   &87.5   &91.5   &86.3   &\textbf{81.7}   &\textbf{73.8}   &\textbf{80.6}   &\textbf{72.9}   &\textbf{84.0}   &\textbf{76.7}
\\ \bottomrule
\end{tabular}
\label{table:seg_results}
\end{table*}
% ****************** Table 1: seg_results ******************

% ****************** Table 2: det_results ******************
\begin{table*}[t]
\centering
\caption{Comparisons of different settings applied on three polyp detection baselines.}
\renewcommand\tabcolsep{3.4pt}
\begin{tabular}{@{}lcccccccccc|cc@{}}
\toprule
\multirow{2}{*}{Methods} 
& \multicolumn{2}{c}{EndoScene}                      
& \multicolumn{2}{c}{ClinicDB}     
& \multicolumn{2}{c}{Kvasir}   
& \multicolumn{2}{c}{ColonDB}   
& \multicolumn{2}{c}{ETIS}   
& \multicolumn{2}{|c}{Overall}  
\\ \cmidrule(l){2-13} 
&AP  &F1  &AP  &F1  &AP  &F1  &AP  &F1  &AP  &F1  &AP  &F1        \\             
\midrule
Center.  &86.9   &\textbf{91.4}  &84.7           &89.2   &75.6   &81.4   &62.2   &72.3   &62.7   &70.1   &56.6   &76.0    \\    
+LDM     &84.1   &84.4           &\textbf{90.4}  &89.9   &81.3   &81.8   &73.4   &74.5   &65.2   &71.7   &62.0   &76.9 \\
+SDM     &\textbf{87.8}          &86.9  &88.7    &\textbf{91.0}  &77.0   &82.8   &71.8   &78.1   &68.2   &72.6   &61.8   &79.1 \\
+\textbf{Ours}  &85.0  &89.1  &86.1  &90.8  &\textbf{77.3}  &\textbf{84.7}  &\textbf{74.2}  &\textbf{80.2}  &\textbf{68.7}  &\textbf{75.6}  &\textbf{65.7}  &\textbf{81.3} \\
\midrule
Sparse.  &89.9   &87.8   &81.4   &86.4   &75.6   &80.2   &78.2   &73.2   &63.8   &62.4   &63.7   &73.2 \\
+LDM     &87.4   &76.3   &\textbf{95.0}   &\textbf{93.5}   &81.5   &58.8   &80.0   &71.0   &64.4   &54.3   &65.3   &66.3\\
+SDM     &\textbf{94.5}   &\textbf{90.5}   &88.7   &86.5   &79.0   &80.5   &\textbf{81.4}   &76.8   &67.8   &67.1   &65.2   &76.7\\
+\textbf{Ours}     &92.8   &86.2   &92.2   &90.6   &\textbf{81.6}   &\textbf{82.3}   &80.1   &\textbf{79.8}   &\textbf{72.4}   &\textbf{70.4}   &\textbf{66.4}   &\textbf{79.0} \\
\midrule
Deform.  &\textbf{98.1}   &\textbf{94.4}   &89.7   &89.9   &\textbf{80.2}   &74.4   &\textbf{82.2}   &75.5   &65.3   &54.7   &64.5   &71.8 \\
+LDM     &94.6   &90.5   &91.6   &89.5   &79.3   &73.4   &78.0   &73.2   &69.0   &64.0   &63.4   &73.3\\
+SDM     &96.0   &90.6   &90.3   &91.2   &82.2   &78.9   &80.1   &75.1   &67.5   &66.7   &65.1   &75.8\\
+\textbf{Ours}     &94.7   &94.3   &\textbf{92.3}   &\textbf{92.0}   &80.0   &\textbf{80.3}   &81.4   &\textbf{77.3}   &\textbf{74.1}   &\textbf{69.3}   &\textbf{67.9}   &\textbf{77.9}
\\ \bottomrule
\end{tabular}
\label{table:det_results}
\end{table*}
% ****************** Table 2: det_results ******************

\subsection{ArSDM Experimental Settings}
We conducted our experiments on five public polyp segmentation datasets: EndoScene~\cite{vazquez2017benchmark}, CVC-ClincDB/CVC-612~\cite{bernal2015wm}, CVC-ColonDB~\cite{tajbakhsh2015automated}, ETIS~\cite{silva2014toward} and Kvasir~\cite{jha2020kvasir}. Following the standard of PraNet, 1,450 image-mask pairs from Kvasir and CVC-ClinicDB are taken as the training set. The evaluations are conducted on the five datasets separately to verify the learning and generalization capability. The training image-mask pairs are padded to have the same height and width and then resized to the size of $384 \times 384$. Experiments with prediction-guided sample refinement are trained with around one-half NVIDIA A100 days, while others are trained with approximately one day for convergence. We use the DDIM sampler with a maximum timestep of 200 for sampling images.

\subsection{Downstream Experimental Settings}
We conduct the evaluation of our methods and the state-of-the-art counterparts on polyp segmentation and detection tasks. 
We generated the same number of samples as the diffusion training set using the original masks, and then combined them to create a new downstream training set. We employed PraNet~\cite{fan2020pranet}, SANet~\cite{wei2021shallow}, and Polyp-PVT~\cite{dong2021PolypPVT} as baseline segmentation models with default settings, and evaluated them using mean Intersection over Union (IoU) and mean Dice metrics. For detection, we selected CenterNet~\cite{zhou2019objects}, Sparse-RCNN~\cite{sun2021sparse}, and Deformable-DETR~\cite{zhu2020deformable} as baseline models with the same settings as the original papers, and evaluated them using Average Precision (AP) and F1-scores.

\subsection{Quantitative Comparisons}
The experimental results presented in Table \ref{table:seg_results} and \ref{table:det_results} demonstrate the effectiveness of our proposed method in training better downstream models to achieve superior performance. Specifically, data generated by our approach assists the significant improvements for each model in mDice and mIoU, with increases of 6.0\% and 5.7\% over PraNet, 2.1\% and 2.7\% over SANet, and 0.7\% and 0.7\% over Polyp-PVT. We also observe superior AP and F1-scores compared to CenterNet, Sparse-RCNN, and Deformable-DETR trained with original data, with gains of 9.1\% and 5.3\%, 2.7\% and 5.8\%, and 3.4\% and 6.1\%, respectively.
Moreover, we conducted a comprehensive comparison with SOTA models, noting that these models were not specifically designed for colonoscopy images and may generate data that hinder the training process or lack the ability for effective improvement. Nevertheless, our experimental results confirm the superiority of our proposed method.

\noindent\textbf{Ablation Study}~
% ****************** Table 3: ablation_study ******************
\begin{figure}[t]
\centering
\begin{minipage}{\textwidth}
\centering
\begin{minipage}[t]{0.51\textwidth}
\makeatletter\def\@captype{table}
\centering
\caption{Ablation study of different com-\\ponents on polyp segmentation tasks.}
\renewcommand\tabcolsep{3pt}
\centering
\begin{tabular}{@{}cccccc@{}}
\toprule
\multicolumn{2}{c}{Methods} & \multicolumn{2}{c}{PraNet}  & \multicolumn{2}{c}{SANet} \\
Ada.   & Ref.   & mDice          & mIoU       & mDice         & mIoU            \\ \midrule
\xmark & \xmark & 76.4           & 69.6       & 80.4          & 72.9                  \\
\cmark & \xmark & 79.1           & 72.4       & 80.5          & 72.8                  \\
\xmark & \cmark & 78.5           & 71.5       & 81.1          & 73.2              \\
\cmark & \cmark & \textbf{80.0} & \textbf{73.2} & \textbf{81.5} & \textbf{74.1}    \\ \bottomrule
\end{tabular}
\label{table:ablation_seg}
\end{minipage}
\begin{minipage}[t]{0.46\textwidth}
\makeatletter\def\@captype{table}
\centering
\caption{Ablation study of different components on polyp detection tasks.}
\renewcommand\tabcolsep{4pt}
\centering
\begin{tabular}{@{}cccccc@{}}
\toprule
\multicolumn{2}{c}{Methods} & \multicolumn{2}{c}{CenterNet}  & \multicolumn{2}{c}{Sparse.} \\
Ada.   & Ref.   & AP            & F1        & AP            & F1            \\ \midrule
\xmark & \xmark & 61.8          & 79.1      & 65.2          & 76.7                  \\
\cmark & \xmark & 62.2          & 80.1      & 65.8          & 77.2                  \\
\xmark & \cmark & 64.0          & 80.4      & 66.0          & 77.6              \\
\cmark & \cmark & \textbf{65.7} & \textbf{81.3} & \textbf{66.4} & \textbf{79.0}    \\ \bottomrule
\end{tabular}
\label{table:ablation_det}
\end{minipage}
\end{minipage}
\end{figure}
% ****************** Table 3: ablation_study ******************
We conducted an ablation study to assess the importance of each proposed component. Table~\ref{table:ablation_seg} and \ref{table:ablation_det} report the overall accuracies on the test set. The results demonstrate both components contribute to the accuracy improvement of baseline models, indicating their essential roles in achieving the best final performance.

% ****************** Figure 3: samples ******************
\begin{figure}[t]
    \centering
    \includegraphics[width=\textwidth]{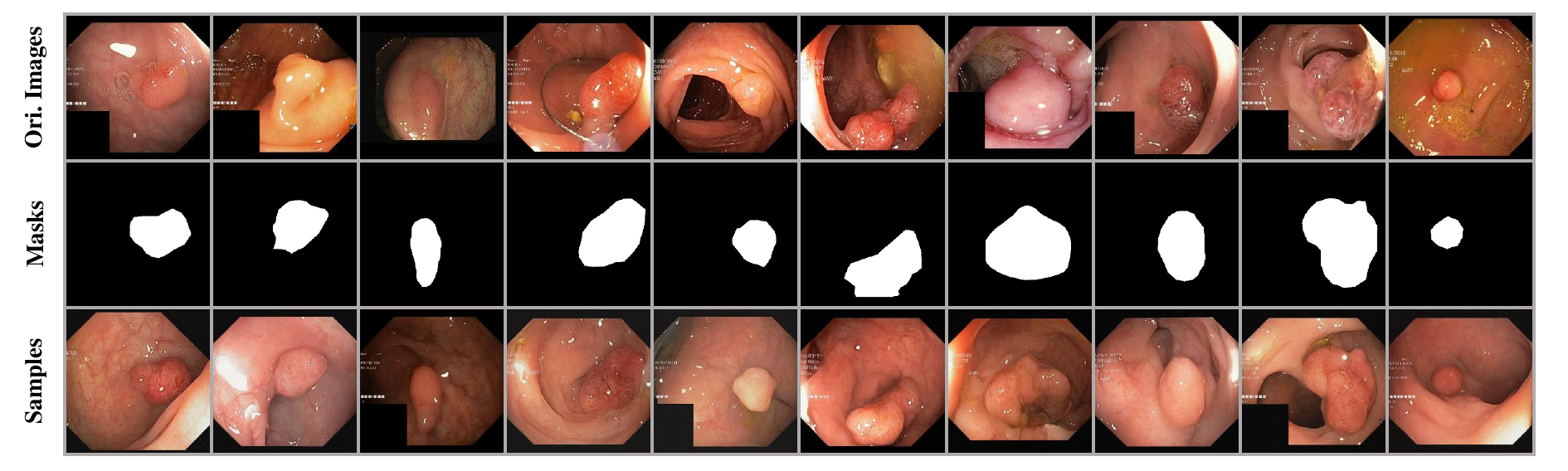}
    \caption{Illustration of generated samples with the corresponding masks and original images for comparison reference.} 
    \label{fig:samples}
\end{figure}
% ****************** Figure 3: samples ******************
\subsection{Qualitative Analyses}
To further investigate the generative performance of our approach, we present visualization results in Fig.~\ref{fig:samples}, which displays the generated samples and their corresponding masks, alongside the original images for reference. The generated samples demonstrate differences from the original images in both the polyp regions and the backgrounds while maintaining alignment with the masks. Additionally, we sought evaluations from medical professionals to assess the authenticity of the generated samples, and non-medical professionals to locate polyps in the images, which yielded positive feedback on the quality of the generated samples.

% ****************** Section 4 ******************
\section{Conclusion}
Automatic generation of annotated data is essential for colonoscopy image analysis, where the scale of existing datasets is limited by the expertise and time required for manual annotation. In this paper, we propose an adaptive refinement semantic diffusion model (ArSDM) for generating colonoscopy images while preserving annotations by introducing innovative adaptive loss re-weighting and prediction-guided sample refinement mechanisms. To evaluate our approach comprehensively, we conduct polyp segmentation and detection experiments on five widely used datasets, where experimental results demonstrate the effectiveness of our approach, in which model performances are greatly enhanced with little synthesized data.

% ****************** Section 5 ******************
\section*{Acknowledgement}
This work was supported in part by the Chinese Key-Area Research and Development Program of Guangdong Province (2020B0101350001), in part by the Shenzhen General Program (No.~JCYJ20220530143600001), in part by the National Natural Science Foundation of China (NO.~61976250), in part by the Shenzhen-Hong Kong Joint Funding (No.~SGDX20211123112401002), in part by the Shenzhen Science and Technology Program (NO.~JCYJ20220818103001002, NO.~JCYJ20220530141211024), and in part by the Guangdong Provincial Key Laboratory of Big Data Computing, The Chinese University of Hong Kong, Shenzhen.

\bibliographystyle{splncs04}
\bibliography{paper1249}
\end{document}